\documentclass[letterpaper, 10pt, conference]{ieeeconf}
\usepackage{times}
\usepackage[pdftex]{graphicx}

\usepackage{amsmath,amssymb,amsopn,amstext,amsfonts, amsthm}
\usepackage{cancel}
\usepackage[space]{cite}
\usepackage{pdfsync}
\usepackage{balance}
\usepackage{color}
\usepackage{mathtools}
\usepackage[ruled,norelsize, linesnumbered]{algorithm2e}
\usepackage{bm}
\usepackage{diagbox}
\usepackage{float}
\usepackage[outdir=./]{epstopdf}
\usepackage{url}
\usepackage{multirow}
\usepackage{makecell}
\usepackage{adjustbox}
\usepackage{tabularx, booktabs}
\let\labelindent\relax
\usepackage{enumitem}
\usepackage[font=small,skip=2pt]{caption}
\usepackage[linkcolor=black,citecolor=black,urlcolor=black,colorlinks=true]{hyperref}
\usepackage{subcaption}
\usepackage{bookmark}

\theoremstyle{remark}

\newcommand{\func}{\mathtt}

\SetKw{Continue}{continue}
\makeatletter
\newcommand{\removelatexerror}{\let\@latex@error\@gobble}
\makeatother

\DeclareGraphicsExtensions{.pdf,.png,.jpg,.eps}
\IEEEoverridecommandlockouts
\title{\LARGE \bf FlowMap: Path Generation for Automated Vehicles in Open Space Using Traffic Flow}
\author{Wenchao Ding$^1$, Jieru Zhao$^{2\dagger}$, Yubin Chu$^3$, Haihui Huang$^3$, Tong Qin$^3$, \\
Chunjing Xu$^3$, Yuxiang Guan$^1$ and Zhongxue Gan$^1$%
\thanks{\noindent Corresponding author: Jieru Zhao. $^1$Academy for Engineering and Technology, Fudan University, China. $^2$Dept. of Computer Science and Engineering, Shanghai Jiao Tong University, China. $^3$IAS BU Smart Driving Product Dept, Huawei Technologies, China. This work is partially sponsored by the National Natural Science Foundation of China (62102249) and Shanghai Pujiang Program (21PJ1408200). Video link \url{https://www.youtube.com/watch?v=GJ_Fm1Iu8Nc}. \newline
{\tt\small Email: dingwenchao@fudan.edu.cn}%
}}
\begin{document}

\maketitle
\thispagestyle{empty}
\pagestyle{empty}

\begin{abstract}
	 There is extensive literature on perceiving road structures by fusing various sensor inputs such as lidar point clouds and camera images using deep neural nets. Leveraging the latest advance of neural architects (such as transformers) and bird-eye-view (BEV) representation, the road cognition accuracy keeps improving. However, how to cognize the ``road'' for automated vehicles where there is no well-defined ``roads'' remains an open problem. For example, how to find paths inside intersections without HD maps is hard since there is neither an explicit definition for ``roads'' nor explicit features such as lane markings. The idea of this paper comes from a proverb: it becomes a way when people walk on it. Although there are no ``roads'' from sensor readings, there are ``roads'' from tracks of other vehicles. In this paper, we propose FlowMap, a path generation framework for automated vehicles based on traffic flows. FlowMap is built by extending our previous work RoadMap~\cite{qin2021roadmap}, a light-weight semantic map, with an additional traffic flow layer. A path generation algorithm on traffic flow fields (TFFs) is proposed to generate human-like paths. The proposed framework is validated using real-world driving data and is amenable to generating paths for super complicated intersections without using HD maps.
\end{abstract}

\section{Introduction}\label{sec:introduction}
To achieve autonomous driving in city environments, high-definition maps (HD maps) are a common and viable solution thanks to their abundant semantic and geometric information. However, HD map is hard to maintain at scale with affordable costs~\cite{casas2021mp3}. Automated driving without HD maps, i.e., mapless autonomous driving, is gaining excessive attention nowadays. There is extensive literature on perceiving map elements online using deep neural nets~\cite{li2022bevformer,qili2022hdmapnet,liu2022bevfusion,liu2022vectormapnet}. The accuracy of road cognition continues to improve thanks to the latest advances of neural network architects like transformers and bird-eye-view (BEV) representation. However, it is difficult to perceive ``roads'' when there are no features (such as lane markings) from sensor readings. For example, finding drivable paths inside a big intersection without HD maps is a challenging task~\cite{tesla2021}.

\begin{figure*}[t]
	\centering
	\hspace*{\fill}
	\begin{subfigure}[b]{0.24\textwidth}
		\centering
		\includegraphics[width =0.98\textwidth]{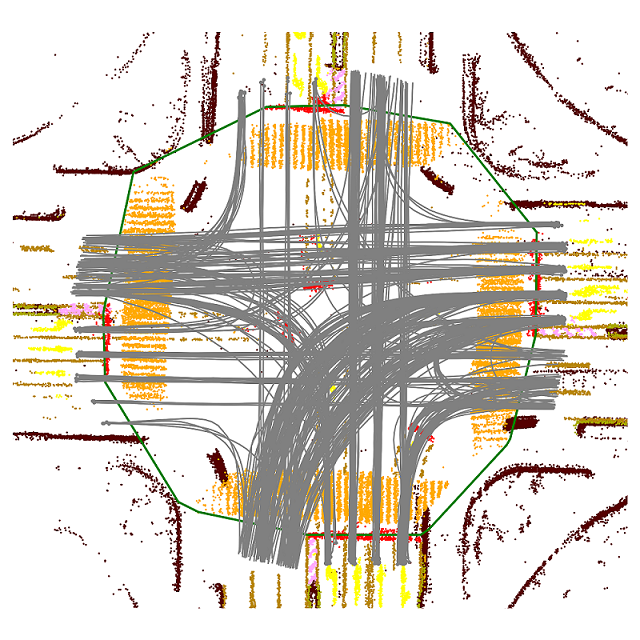}
		\caption{\small{Complete traffic flows}}\label{fig:2_1}
	\end{subfigure}
	\hspace*{\fill}
	\begin{subfigure}[b]{0.24\textwidth}
		\centering
		\includegraphics[width =0.98\textwidth]{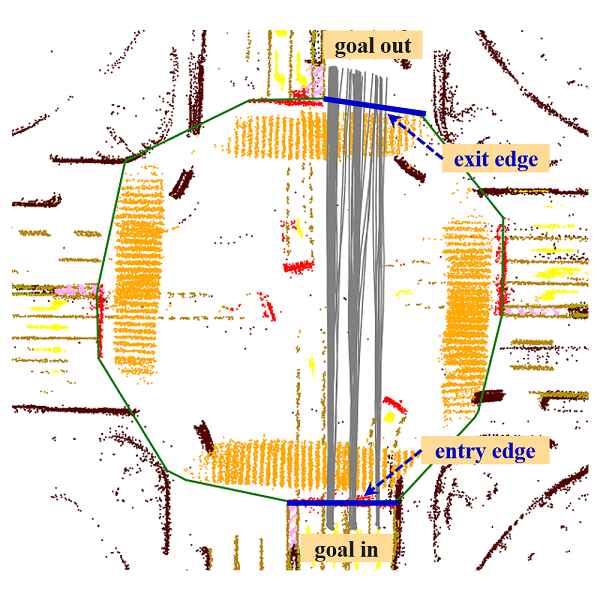}
		\caption{\small{Flows from a particular goal}}\label{fig:2_2}
	\end{subfigure}
	\begin{subfigure}[b]{0.24\textwidth}
		\centering
		\includegraphics[width =0.98\textwidth]{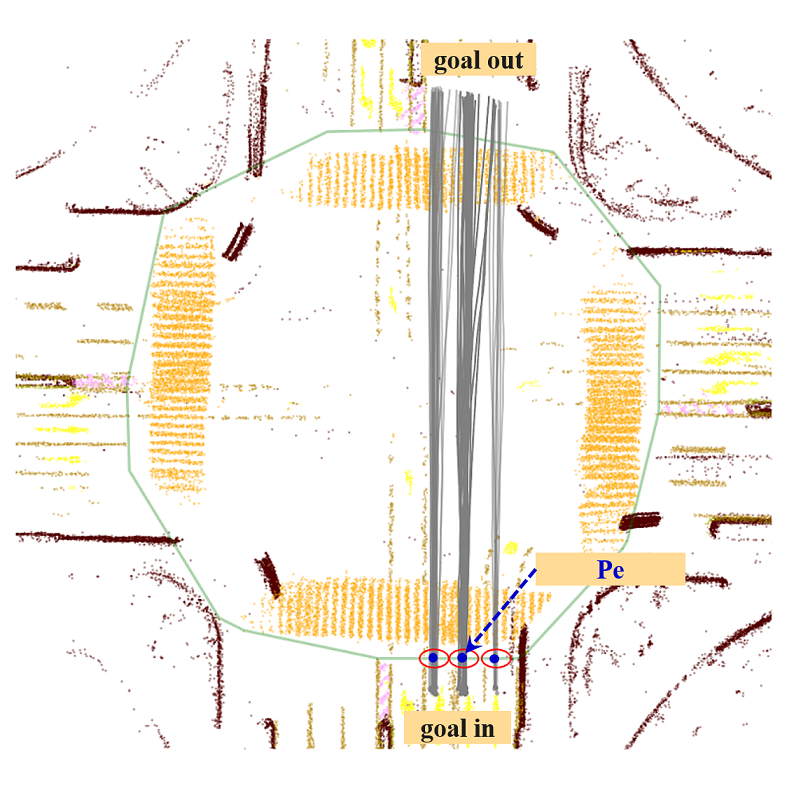}
		\caption{\small{Entry point clustering}}\label{fig:2_3}
	\end{subfigure}
	\begin{subfigure}[b]{0.24\textwidth}
		\centering
		\includegraphics[width =0.98\textwidth]{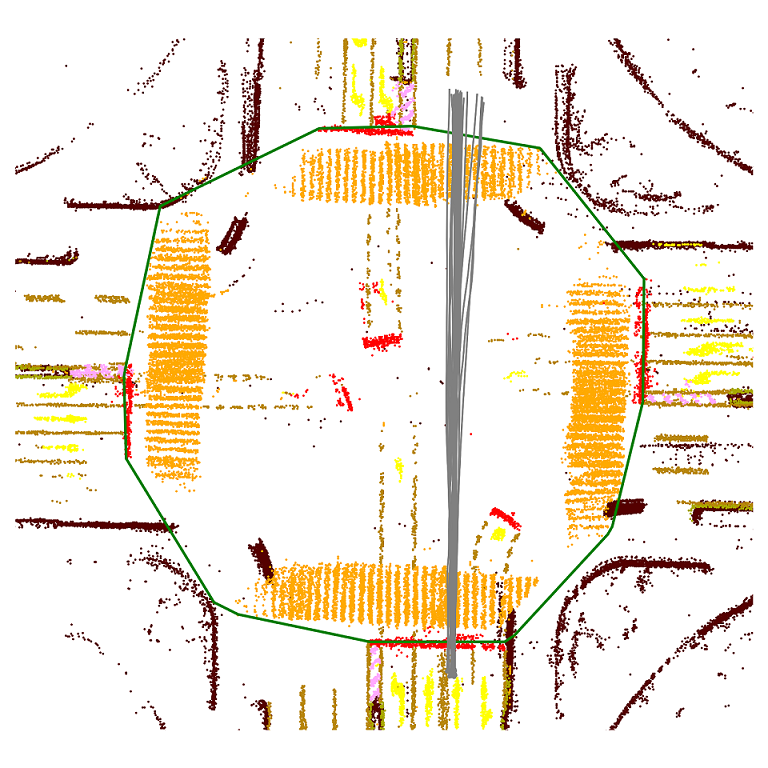}
		\caption{\small{Flow from an entry point}}\label{fig:2_4}
	\end{subfigure}		
	\caption{An illustration of the preprocessing of traffic flows. Raw traffic flows are depicted in (a) and the flows from different directions mess up. First, traffic flows are grouped by entry goal $g_{\text{in}}$ and exit goal $g_{\text{out}}$ (b). Second, entry point clustering is carried out and $3$ entry points $p_e$ are found (c). Third, flows for a particular $p_e$ are shown and form one channel of $G$ (d).}\label{fig:flow_pre}
\end{figure*}

Human drivers do not drive solely based on lane or road markings. When human drivers come to unknown places, they tend to follow the tracks of other human drivers. Similar to the proverb ``it becomes a way when people walk on it,'' it becomes a lane when people drive on it. However, simply following the leading vehicle is cumbersome since there is no global topological understanding of the environment. There is much more information to be mined from daily point-to-point driving. For example, suppose there is a big intersection in your path of everyday driving. As you pass the intersection several times, \textit{traffic flows} of the intersection can be accumulated as shown in Fig.~\ref{fig:2_1}. The traffic flows consist of a number of vehicle trajectories which are produced by an onboard tracking module. Traffic flows are beneficial for path planning since they contain abundant driving behaviors.

\begin{figure}[t]
	\vspace{+0.3cm}
	\centering
	\includegraphics[width=0.48\textwidth]{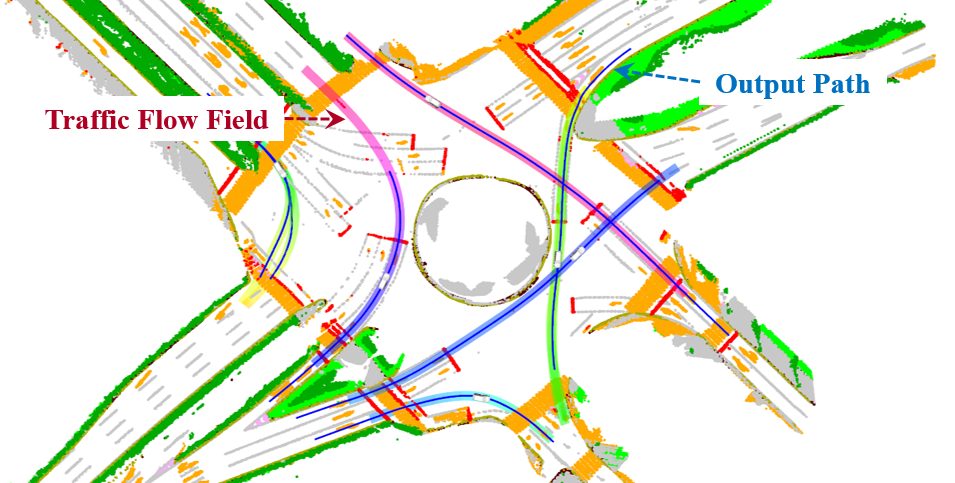}
	\caption{An illustration of FlowMap in a super large intersection. The semantic map produced by our previous work RoadMap~\cite{qin2021roadmap} is illustrated, where lane markings (\textit{gray}), stop lines (\textit{red}), cross walks, and turn markings (\textit{orange}) are stored in semantic point clouds. Traffic flow fields are shown using color gradients, and the color represents the flow direction. The path generated by FlowMap is shown in \textit{blue} lines. With the help of the traffic flow field, human-like driving paths can be elegantly extracted.}\label{fig:cover}
	\vspace{-0.5cm}
\end{figure}

RoadMap~\cite{qin2021roadmap}, our previous work, provides a systematic solution for producing lightweight semantic maps~\cite{jeong2017road,qin2020avp,herb2019crowd}. Thanks to RoadMap, traffic flows are easy to obtain. They can be obtained by extending RoadMap with an additional traffic flow layer. The traffic flow layer mainly contains social vehicle trajectories from onboard tracking. The key point is how to properly manage, organize, and utilize traffic flow. In this paper, we propose using traffic flow fields (TFFs) to model the accumulation and update of traffic flows. The TFF is a discretized grid map synthesized by calculating flow density and direction for each cell. The grid map consists of multiple channels, with each channel representing the flow originating from a particular entry point. The multi-channel design ensures that the flows from different directions do not mess up with each other. Both the semantic map and TFF only contain scalar information and can be easily maintained.

The proposed traffic flow field representation contains considerable information about multi-modal human driving behaviors, but it is unstructured. To facilitate planning and control, we proposed a path generation algorithm to extract human-like driving paths from traffic flow fields. Based on this algorithm, guidance paths can be produced even in large open spaces where there are few road markings.

To summarize, we present FlowMap, a framework for path generation in open space using traffic flow. FlowMap mainly consists of a traffic flow field representation and a path generation algorithm built on top of that. FlowMap is amenable to producing guidance paths even in large open spaces and enables human-like driving in such scenarios without HD maps. The major contributions are summarized as follows:
\begin{itemize}
	\item FlowMap, a lightweight framework for path generation in open space without using HD maps.
	\item A traffic flow field representation for modeling driving behaviors from traffic flows.
	\item A path generation algorithm producing human-like guidance paths in open space based on the traffic flow fields. 
	\item Comprehensive analysis of the overall framework in real-world complicated scenarios.
\end{itemize}

The remainder of this paper is organized as follows. The related work is reviewed in Section~\ref{sec:related_work}. An overview of the proposed framework is provided in Section~\ref{sec:system_overview}. The methodology and implementation are detailed in Section~\ref{sec:method} and Section~\ref{sec:implementation}, respectively.
Experimental results and benchmark analysis are elaborated in Section~\ref{sec:experimental_results}. Finally, this paper is concluded in Section~\ref{sec:conclusion}.

\section{Related Work}\label{sec:related_work}
There is extensive literature on road cognition and path planning for automated vehicles. The road cognition problem is first formulated by the 2D detection of lane markings in the image space~\cite{pan2018spatio,neven2018towards}. To improve lane structure preservation during 2D detection, an EL-GAN-based framework is proposed in \cite{ghafoorian2018gan}. Philion proposed FastDraw~\cite{jonah2022fastdraw}, a fully convolutional model of lane detection that learns to decode lane structures in perspective views instead of delegating structure inference to post-processing. After 2D detection, depth information is required to reproject the 2D lane detection results to 3D and obtain drivable paths. Lane markings are critical for 2D detection-based methods and greatly affect the accuracy of lane detection results. For open space where there are no lane markings, the lane detection problem is ill-posed.

\begin{figure}[t]
	\centering
	\vspace{+0.5cm}
	\includegraphics[width=0.48\textwidth]{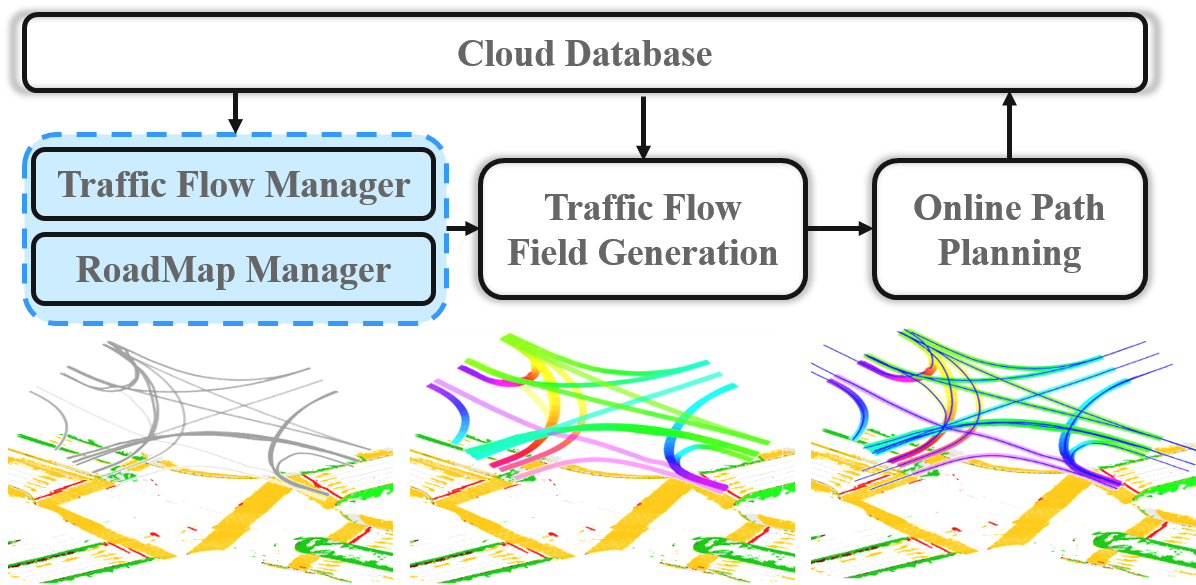}
	\caption{An illustration of the FlowMap framework. First, RoadMap is extended with an additional traffic flow manager, and traffic flows are stored together with the semantic map (left). Second, traffic flow fields are synthesized into traffic flow fields, which is shown using color gradients (middle). Third, online path search is carried out based on semantic map and traffic flow fields (right).}\label{fig:overview}
	\vspace{-0.6cm}
\end{figure}

In recent years, the advances of neural architects such as transformers~\cite{devlin2018bert,vaswani2017attention,carion2022detr} have pushed the limit of road cognition performance forward. The trend is to model the lane prediction problem directly in the form of bird-eye-view (BEV) representation and use multi-head attention to properly find corresponding features from perspective views or lidar point clouds. In~\cite{qili2022hdmapnet}, a semantic map learning method, named HDMapNet, is proposed. HDMapNet encodes image features from surrounding cameras and/or point clouds from LiDAR and predicts vectorized map elements in BEV. In~\cite{li2022bevformer}, a transformer-based semantic learning method is proposed to detect objects and segment road structures simutaneously. The road cognition problem for areas where there are explicit features is solved much better than the original 2D formulation. However, it remains an open problem of how to find ``roads'' where there are no explicit features for neural networks to extract.

Finding ``roads'' in open space is inherently a planning problem. To be specific, the road cognition problem can be modeled as a path searching or optimization problem under the constraints of detected road topology and obstacles. The output path should be collision-free and follow the perceived curbs and boundaries. Hybrid A$^\ast$ search~\cite{dolgov2009hybrid,dolgov2010path} on a state lattice~\cite{mcnaughton2011motion,ziegler2009spatiotemporal} is a popular option for path searching, which can be traced back to DARPA Urban Challenge 2007. Typically, an optimization-based method like elastic bands~\cite{ding2018trajectory,xu2012optimization} is incorporated to compensate for the discretization of the state lattice and improve the overall smoothness. However, collision-free is not the only requirement for a drivable path of automated vehicles. To achieve socially-compliant autonomous driving, human likeness is an important constraint, but is not paid much attention to. Traffic flow is the key to human likeness since it contains useful information of how other vehicles drive.

This paper is motivated by the proverb ``it becomes a way when people walk on it''. To achieve human-like automated driving, it is essential to extract the driving patterns of other vehicles. This problem is not paid much attention to since HD maps are very popular. From the HD maps, reference paths are easy to access. However, how to maintain and update HD map efficiently in an affordable way is still an open question. When there is no HD map, how other vehicles drive is the key to human-like driving.
This paper proposed a novel method to incorporate traffic flow to achieve human-like path generation in open space. Traffic flow is similar to occupancy grids and occupancy flow~\cite{mahjourian2022occupancy,mahjourian2019discrete,hong2019rules} to some extent. However, occupancy flow is designed for short-term behavior prediction, while FlowMap is designed for generating long-term drivable paths.

\section{Framework Overview}\label{sec:system_overview}
An overview of FlowMap is depicted in Fig.~\ref{fig:overview}. The framework consists of three major components: a lightweight semantic mapping module with an additional traffic flow management function, a traffic flow field generation module, and an online path planning and smoothing module. The overall system not only deals with the problem of how to maintain the traffic flow in an affordable way but also provides algorithms and methods to support offline traffic flow processing and online path generation.

The FlowMap framework is built on top of our previous work, RoadMap. RoadMap provides a lightweight semantic mapping framework. It takes sparse semantic point clouds from vehicles as input and fuses them into a global lightweight semantic map. Apart from semantic point clouds, we add a traffic flow management function to RoadMap, which essentially maintains onboard detection and tracking results together with semantic point clouds. Traffic flows and semantic points are then fused in the cloud, where traffic flows are further processed in the form of traffic flow fields, which facilitate path planning. In the online process which is triggered when a particular automated vehicle enters a region of interest (ROI), the vehicle is amenable to generating a human-like driving path with the help of the traffic flow fields from the cloud. The path generation process takes the trend of traffic flow, the semantic map, and driving comforts into consideration simutaneously.

\section{Path Generation Using Traffic Flow}\label{sec:method}
In this section, we first introduce the formulation of the traffic flow and techniques to acquire traffic flows. Then we elaborate on how to transform traffic flows into traffic flow fields, which facilitates the online planning process. Finally, we present the online planning process, including an online path search stage and a following path optimization stage.

\subsection{Traffic Flow}\label{sec:trafficflow_acq}
The input of traffic flow comes from onboard vehicle tracking modules. Denote $b^{t}_{i} = (p, l, w)$ as the tracking result for a specific tracked vehicle with id $i$ at time stamp $t$, where $p=(x, y, \theta)$ denotes the tracked bounding box's center coordinates and heading, and $l$, $w$ denote the length and width of the bounding box, respectively. Let $b_{v}$ represent the path of vehicle $v$ across all time stamps tracked. Denote $\mathcal{F} = \{b_{v}|v\in \mathcal{V}\}$ as the set of all observed vehicle traces, and denote $\mathcal{V}$ as the set of vehicle tracking id. Note that the traces can be obtained by a single vehicle or a fleet of vehicles at different space and time. An example of $\mathcal{F}$ at a particular intersection is shown in Fig.~\ref{fig:2_1}.

The traffic flow generation is lightweight and only relies on onboard tracking. However, there are three major problems. First, many observed trajectories have very short lifetimes due to occlusion. Second, perception errors exist and may induce abnormal trajectories, such as trajectories drifting or colliding with obstacles. Third, there are stationary or short trajectories which contain limited information.

To this end, filtering is required to clean $\mathcal{F}$ so that it only contains high-quality vehicle traces. The filtering process is implemented based on four criteria: the lifetime of the trace; the total travel distance of the trace; the existence of abnormal drifting; the collision with obstacles; and we only keep trajectories that have a clear entry point and exit point with respect to the region of interest. After filtering, each trace $b_{v}$ is marked with meta information $m_{v}=(l_{\text{entry}},l_{\text{exit}}, p_{\text{start}})$, where $l_{\text{entry}}$ and $l_{\text{exit}}$ denote the corresponding edges of ROI (which is parameterized as a polgon in RoadMap) which intersect with the trajectory, and $p_{\text{start}}$ denotes the intersection point of $l_{\text{entry}}$. We denote the traffic flow with extra meta information as $\mathcal{F^{'}} = \{(b_{v}, m_{v})|v\in\mathcal{V}\}$.
\begin{figure}[t]
	\removelatexerror
	\begin{minipage}{.48\textwidth}
		\begin{algorithm}[H]\label{algo:trafficflow_pre}
			\caption{Preprocessing of traffic flow field}
			Inputs: Traffic flow $\mathcal{F}^{'}$\;
			$\Gamma\leftarrow\emptyset$\;
			$\{\mathcal{F}^{'}_{<g_{\text{in}},g_{\text{out}}>}\}\leftarrow\func{GroupByGoal}(\mathcal{F}^{'})$\;
			\ForEach{$\mathcal{F}^{'}_{<g_{\text{in}},g_{\text{out}}>}\in\{\mathcal{F}^{'}_{<g_{\text{in}},g_{\text{out}}>}\}$}
			{
				$\{p_{e}\}\leftarrow\func{EntryPointClustering}(\mathcal{F}^{'}_{<g_{\text{in}},g_{\text{out}}>}) $\;
				$\{\mathcal{F}^{'}_{<g_{\text{in}},g_{\text{out}},p_e>}\}\leftarrow\func{GroupByEntryPoint}(\mathcal{F}^{'},\{p_{e}\})$\;
				$\Gamma\leftarrow\Gamma\cup\{\mathcal{F}^{'}_{<g_{\text{in}},g_{\text{out}},p_e>}\}$\;				
			}
			\Return{$\Gamma$};
		\end{algorithm}
	\end{minipage}
	\vspace{-0.7cm}
\end{figure}

\subsection{Traffic Flow Field Generation} \label{sec:field_gen}
$\mathcal{F}$ is essentially a set of vehicle trajectories organized by spatial location. It is not straightforward to use $\mathcal{F}$ to conduct path generation since no explicit structure is constructed. To this end, we propose generating a traffic flow field $\mathcal{G}$ based on $\mathcal{F}$, which encapsulates flow density and flow direction information in a multi-channel grid map. However, directly rendering $\mathcal{F}$ on a grid map like the occupancy flow presentation~\cite{mahjourian2022occupancy} is not friendly for planning, since flows originating from different directions will mess up with each other. What is worse, flows from different directions are often imbalanced. 

To overcome this, we preporcess $\mathcal{F}^{'}$ by grouping it by entry goal, exit goal, and entry point. The preprocessing algorithm is introduced in Algorithm~\ref{algo:trafficflow_pre}. First, given the meta information $m_v$ for each trace which contains the entry edge $l_{\text{entry}}$ and exit edge $l_{\text{exit}}$ of the ROI polygon, goal indices $g_{\text{in}}$, $g_{\text{out}}$ are generated based on the edges, and $\mathcal{F}^{'}$ is grouped based on the pair of $g_{\text{in}}$ and $g_{\text{out}}$ (Line 3). Second, for each group, multiple entry points $p_{e}$ are obtained through clustering (Line 5).  $p_{e}$ represents lane-level identification for each trace. Third, the flow is further grouped by the entry points, and $\mathcal{F}^{'}_{<g_{\text{in}},g_{\text{out}},p_e>}$ is obtained. $\mathcal{F}^{'}_{<g_{\text{in}},g_{\text{out}},p_e>}$ can be understand as a subset of flow indexed by the entry ``lane''. The reason for extracting lane-level information is that there is intrinsic multi-modality in the traffic flow, even with the same entry and exit goal. A typical example of the preprocessing is depicted in Fig.~\ref{fig:flow_pre}.

\begin{figure}[t!]
	\centering
	\vspace{+0.3cm}
	\begin{subfigure}[b]{0.23\textwidth}
		\centering
		\includegraphics[width=\textwidth]{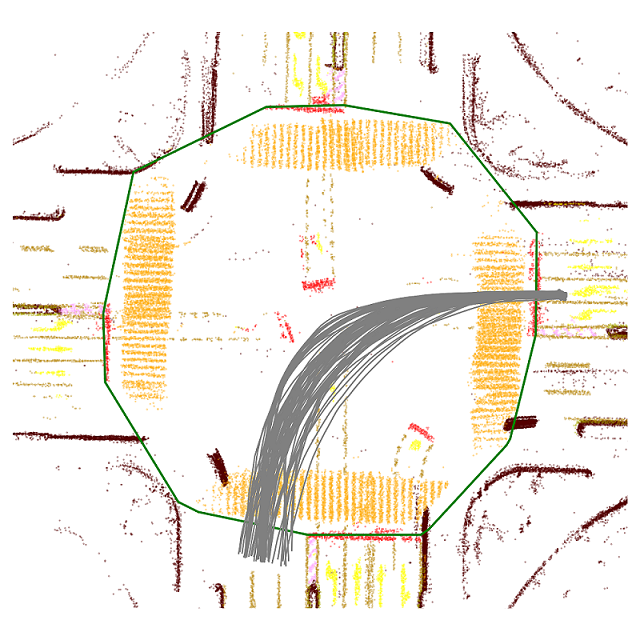}
		\caption[Network2]%
		{{\small Flow for one entry point}}    
		\label{fig:field_gen_flow}
	\end{subfigure}
	\hfill
	\begin{subfigure}[b]{0.23\textwidth}  
		\centering 
		\includegraphics[width=\textwidth]{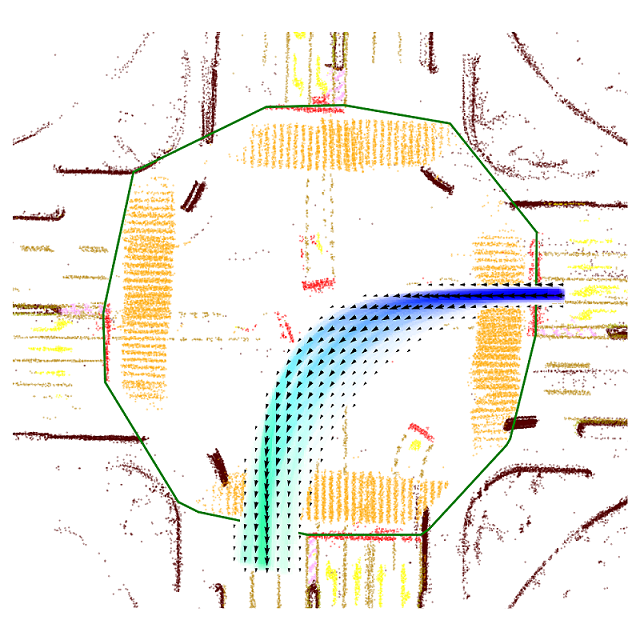}
		\caption[]%
		{{\small Traffic flow field}}    
		\label{fig:field_gen_field}
	\end{subfigure}
	\vskip\baselineskip
	\begin{subfigure}[b]{0.23\textwidth}   
		\centering 
		\includegraphics[width=\textwidth]{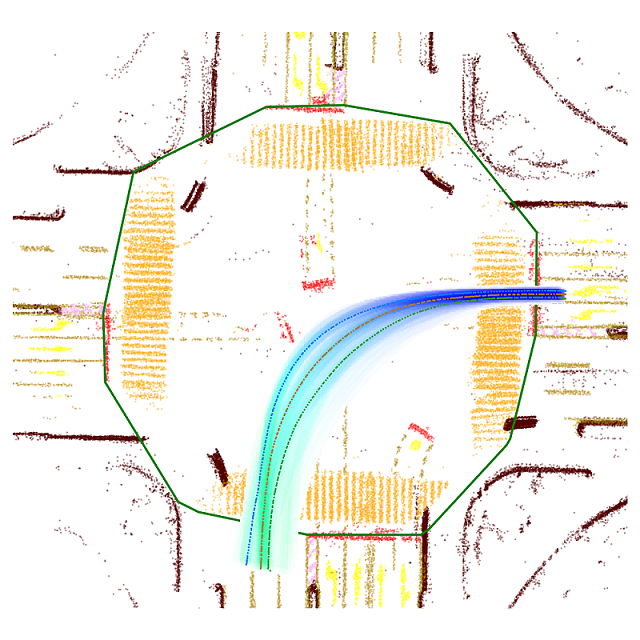}
		\caption[]%
		{{\small Path search}}    
		\label{fig:field_gen_search}
	\end{subfigure}
	\hfill
	\begin{subfigure}[b]{0.23\textwidth}   
		\centering 
		\includegraphics[width=\textwidth]{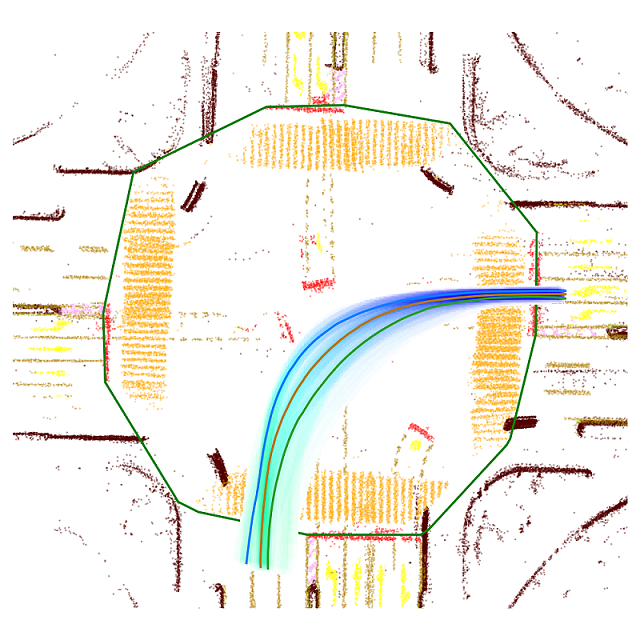}
		\caption[]%
		{{\small Path smoothing}}    
		\label{fig:field_gen_smoothing}
	\end{subfigure}
	\caption[]
	{\small An illustration of the traffic flow field generation and path planning process. For each channel, a traffic flow field is synthesized where flow direction is shown in color gradients and small arrows (b). Multiple human-like left-turn driving paths are found by path search (c). It is interesting that human drivers have different driving patterns when taking left turns as in (b) and our algorithm captures that. Path smoothing is conducted to ensure comfort.} \label{fig:field_gen_path_gen}
	\vspace{-0.5cm}
\end{figure}

Thanks to Algorithm~\ref{algo:trafficflow_pre}, the traffic flows are now nicely organized. We design $\mathcal{G}$ to have a multi-channel structure with each channel representing the flows from a particular entry point $\mathcal{F}^{'}_{<g_{\text{in}},g_{\text{out}},p_e>}$. For each channel, we construct a 2D grid, where each cell contains a tuple $f_{i,j} = (d, \delta_i, \delta_j)$. $d$ denotes the density (number of vehicles passing the cell), and $\delta_i$, $\delta_j$ denote the averaged direction at cell $i,j$. A toy example of one channel of $\mathcal{G}$ is illustrated in Fig.~\ref{fig:field_gen_field}.

Multi-channel traffic flow fields pose an elegant structure to encode human-like driving behaviors. We maintain a separate grid for each ROI polygon. Global traffic flow fields are indexed by spatial location, which is maintained together with RoadMap. To enable continuous updating of traffic flow fields, a first-in-first-out (FIFO) queue is associated with each ROI. When new traffic flow is observed, some of the old traffic flow will vanish. Through this mechanism, we ensure that the traffic flow field is updated to date. 

Moreover, once a significant change in traffic flow fields is observed, which is typically caused by road construction and road re-routing, A map update trigger will be generated by the traffic flow field generation process, which is then sent to RoadMap. We find that the traffic-flow-based update trigger is very useful for detecting the change in road course and can serve as an important complement for lightweight mapping. A typical example is given in Fig.~\ref{fig:4_2_xianshibiangeng}. We integrate the traffic flow update and traffic flow field generation processes into RoadMap. The traffic flow update process consumes much less bandwidth than the original semantic map construction process and does not significantly increase the map maintenance cost.

\subsection{Online Path Generation}
In this section, we go over the path search algorithm, which is based on the multi-channel traffic flow field $\mathcal{G}$, as shown in Algorithm~\ref{algo:path_search}. The path search algorithm is conducted per channel (Line 3). For each channel, we first search for an initial guess based on maximum density (Line 5). The initial guess $l_\text{init}$ is then used to construct a frenet frame~\cite{werling2012mp} for later processing. Based on the frenet frame representation, longitudinal stations (Line 6) and lateral clusters for each longitudinal station (Line 8) are sampled. Dynamic programming is carried out based on longitudinal stations and lateral clusters (Line 11), where edge and node costs determined by the quality of match with respect to $\mathcal{G}_z$. Due to space limitations, the detailed description of the dynamic programming is omitted, and we refer interested readers to~\cite{mcnaughton2011motion} for similar implementations. Multiple potential candidate paths are produced by the dynamic programming algorithm, and a non-maximum suppressing process is utilized to remove the nearly duplicate candidate paths. An example of the searching process is depicted in Fig.~\ref{fig:field_gen_search}.

The output of Algorithm~\ref{algo:path_search} consists of paths originating from each entry lane. Due to the discretization of the grid, the path is not smooth enough for control. To this end, we utilize a local path smoothing~\cite{ding2019safe} based on Quadratic Programming (QP) to smooth the path. Note that~\cite{ding2019safe} was designed for trajectory optimization, and we slightly reformulate it to conduct path smoothing. An example of the path smoothing process is given in Fig.~\ref{fig:field_gen_smoothing}.

\section{Implementation Details}\label{sec:implementation}
\subsection{Traffic Flow Preprocessing}
In traffic flow filtering, only vehicle trajectories that have clear entry and exit points with respect to the ROI polygon are kept. To filter trajectories which collide with static obstacles mapped by RoadMap, we first construct a 2D grid based on the semantics provided by RoadMap. The grid is of $0.2$ m resolution, and obstacles are inflated by half vehicle width when filling in the grid.

In the entry point clustering process (Line 5, Algorithm~\ref{algo:trafficflow_pre}), since the ROI polygon is inaccurate for some cases, the vehicle trajectories may not be perpendicular to $l_\text{entry}$. We refine the edges of the ROI polygon to be perpendicular to the averaged flow direction. 
\begin{figure}[t]
	\removelatexerror
	\begin{minipage}{.48\textwidth}
		\begin{algorithm}[H]\label{algo:path_search}
			\caption{Path Generation on Traffic Flow Fields}
			Inputs: multi-channel traffic flow field $\mathcal{G}$ with each channel denoted as $\mathcal{G}_z$\;
			$\mathcal{L}\leftarrow\emptyset$\;
			\ForEach{$G_z\;\text{of}\;\mathcal{G}$}{
				$\mathcal{D}\leftarrow\emptyset$\;
				$l_{\text{init}}\leftarrow\func{InitialGuessSearch}(\mathcal{G}_z)$ \;
				$\{s_i|i=1,...,N\}\leftarrow\func{StationSampling}(l_{\text{init}})$\;
				\For(){$i=1,...,N$}{
					$\mathcal{D}_i\leftarrow \func{LateralClustering}(s_i)$\;
					$\mathcal{D}\leftarrow \mathcal{D} \cup \mathcal{D}_i$
				}
				$\mathcal{L}_z\leftarrow\func{DPSearch}(\mathcal{D},\{s_i|i=1,...,N\},\mathcal{G}_z)$\;
				$\mathcal{L}_z\leftarrow\func{NonMaximumSuppress}(\mathcal{L}_z)$\;
				$\mathcal{L}\leftarrow\mathcal{L}\cup\mathcal{L}_z$\;
			}
			\Return{$\mathcal{L}$};
		\end{algorithm}
	\end{minipage}
\vspace{-0.5cm}
\end{figure}

\subsection{Traffic Flow Fields}
The resolution of $\mathcal{G}$ is of $0.2$ m. Every cell keeps track of $f_{i,j} = (d, \delta_i, \delta_j)$. $d$ is calculated by counting the number of vehicle traces that overlap with the cell. Note that bounding boxes instead of center points are used for calculating $d$. For $\delta_i$ and $\delta_j$, they are averaged direction over all the traces which overlap with the cell. The number of channels for $\mathcal{G}$ depends on the total number of entry points clustered. The storage of $\mathcal{G}$ utilizes its sparse structure to save bandwidth.

\subsection{Online Path Generation}
For the initial guess search (Line 5, Algorithm~\ref{algo:path_search}), we start from the entry point for each layer, and use dynamic programming to search for a path which has the best match with respect to $\mathcal{G}_z$. For the initial guess search, we have two weighted costs, namely, the density cost and the direction cost, which punish the paths that enter low-density areas or do not match the field direction. The resolution of station sampling (Line 6) is set to $3$ m. The lateral clustering strategy (Line 8) is similar to entry point clustering. Another dynamic programming search is in Line 11, which is built on the graph formed by the sampled stations and lateral clusters, where $\mathcal{G}_z$ is used for cost evaluation. When conducting non-maximum suppression (Line 12), we first rank the candidate paths by their costs, output by the dynamic programming, and then iterate from the path that has the lowest cost. The suppressing criteria is based on the percentage of stations at which the minimum lateral distance to previous paths is greater than $2$ m. If a path contains over $20\%$ of the stations that have significant lateral differences, the path will be considered as a new candidate.

\begin{figure}[t]
	\centering
	\begin{subfigure}[b]{0.23\textwidth}
		\centering
		\includegraphics[width=\textwidth]{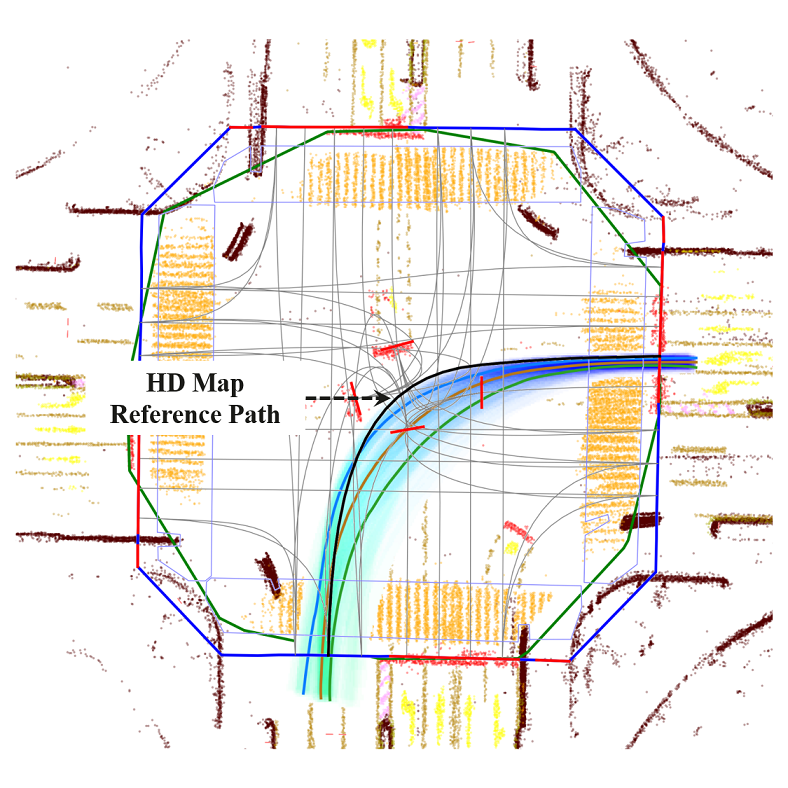}
		\caption[Network2]%
		{{\small Aligning with HD map}}    
		\label{fig:4_1_hd_compare}
	\end{subfigure}
	\hfill
	\begin{subfigure}[b]{0.23\textwidth}  
		\centering 
		\includegraphics[width=\textwidth]{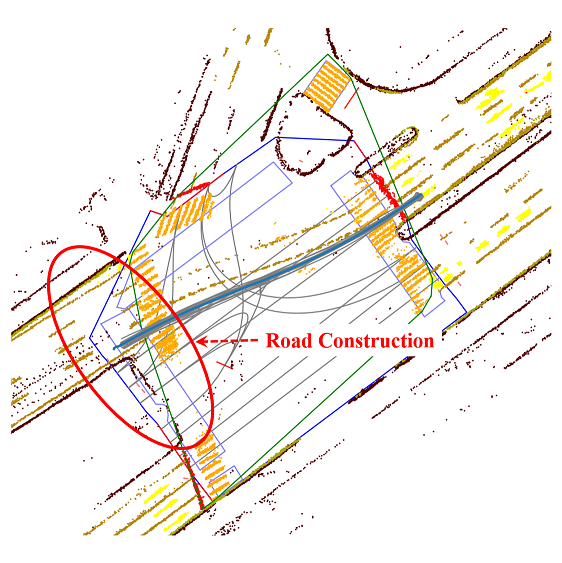}
		\caption[]%
		{{\small Detecting HD map error}}    
		\label{fig:4_2_xianshibiangeng}
	\end{subfigure}
	\caption[]
	{\small An illustration of aligning FlowMap with HD map for quantitative evaluation. Note that the reference left-turn path (\textit{black}) queried from HD map is actually not human-like and differs from the traffic flow (a). By contrast, our FlowMap can produce multiple candidate left-turn paths that match human driving patterns. In (b), the traffic flow does not match the HD map, which identifies the HD map error caused by road construction and road re-routing.} 
	\label{fig:exp_hd}
	\vspace{-0.8cm}
\end{figure}

\section{Experimental Results}\label{sec:experimental_results}
In this section, we present the qualitative and quantitative results for the evaluation of FlowMap. We use real-world datasets for all the experiments. For qualitative analysis, we present typical path generation results for intersections, which are the most common open spaces encountered by automated vehicles. For quantitative analysis, we use two in-house datasets, i.e., \textit{inter-mild} and \textit{inter-hard}. \textit{inter-mild} consists primarily of small cross roads and t-junctions as illustrated in Fig.~\ref{fig:field_gen_path_gen}, while \textit{inter-hard} contains large and complex intersections as shown in Fig.~\ref{fig:exp_qualitative}. The two datasets are visualized in the attached video.
\begin{table*}[!tb]
	\centering
	\vspace{+0.3cm}
	\caption{Comparison with reference paths from HD map.\label{tab:hd_compare}}
	\begin{tabular}{|c||c||c|c|c|c|c|c|c|c|c|c|}
	\hline 
	\multirow{2}{*}{\textbf{Dataset}} & \multirow{2}{*}{\textbf{Type}} & \multicolumn{2}{c|}{ADE} & \multicolumn{2}{c|}{MDE}  & \multicolumn{2}{c|}{DE@5$m$}  & \multicolumn{2}{c|}{DE@35$m$}  &  \multicolumn{2}{c|}{DE@55$m$} \\ \cline{3-12}
									 && avg (m) & std (m) & avg (m) & std (m) & avg (m) & std (m) & avg (m) & std (m) & avg (m) & std (m)\\
	\hline\hline
	\multirow{3}{*}{\textbf{\makecell{Inter\\mild}}}  & Left      & 1.53  & 1.09 & 2.83 & 1.67 & 1.35 & 1.59 & 1.68 & 1.43 & 1.22 & 0.78  \\
	                               					  & Right     & 1.50  & 1.30  & 2.52  & 1.72 & 1.14 & 1.15 & 0.80 & 0.70 & 0.22 & -\\
	                               					  & Straight  & 0.52   & 0.65   & 0.82  & 0.99    & 0.50 & 0.77 & 0.54 & 0.63 & 0.48 & 0.60 \\
    \hline
	\multirow{3}{*}{\textbf{\makecell{Inter\\hard}}}   & Left  & 1.89   & 1.42  & 3.55   &  2.18  & 2.59 & 2.58 & 1.14 & 1.05 & 1.24 & 0.94 \\
	                                                   & Right & 1.27   & 0.93  & 2.49   & 2.01  & 0.92 & 1.02 & 0.57 & 0.71 & 1.06 & -\\
	                               					   & Straight  & 0.57   &  0.47 & 1.04  & 0.95 & 0.45 & 0.58 & 0.58 & 0.57 &- &- \\
    \hline
\end{tabular}
\vspace{-0.3cm}
\end{table*}
\subsection{Qualitative Results}
\subsubsection{Path Generation for Super Large Intersections}
\begin{figure}[t]
	\centering
	\begin{subfigure}[b]{0.45\textwidth}
		\centering
		\includegraphics[width=\textwidth]{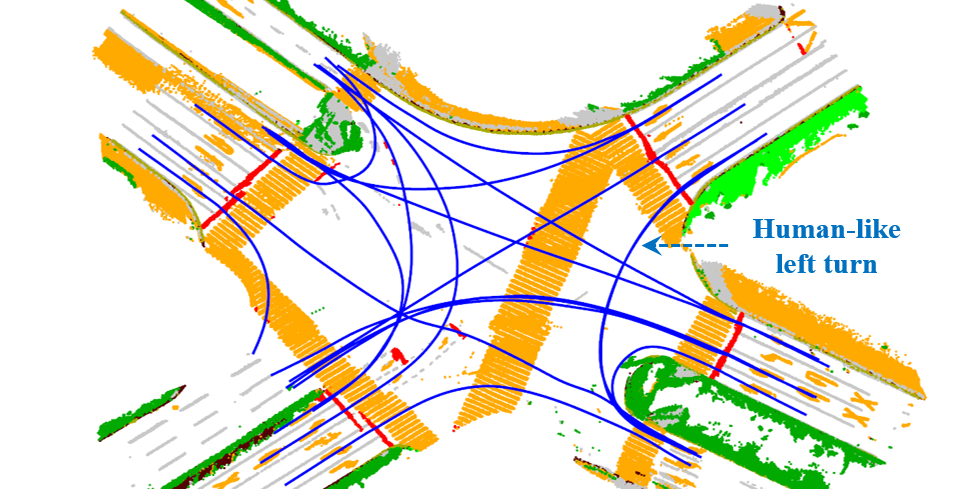}
		\caption[Network2]%
		{{\small Intersection I. Trip count: $17$.}}    
		\label{fig:5_1_lujiazui}
	\end{subfigure}
	\hfill
	\begin{subfigure}[b]{0.45\textwidth}  
		\centering 
		\includegraphics[width=\textwidth]{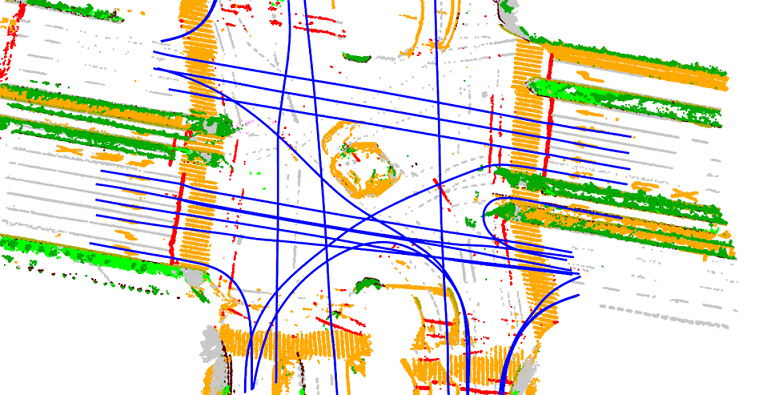}
		\caption[]%
		{{\small Intersection II. Trip count: $27$.}}    
		\label{fig:5_2_huaxiaxi}
	\end{subfigure}
	\caption[]
	{\small An illustration of quantitative results in some super-large intersections. The generated paths are marked in~\textit{blue}. The number of trips passing through the intersection is indicated in the caption. For tens of trips, the observed traffic flows are already enough to construct an abundant topology of large intersections.} 
	\label{fig:exp_qualitative}
	\vspace{-0.8cm}
\end{figure}

In this section, we present challenging scenarios such as super large and irregular intersections, as shown in Fig.~\ref{fig:exp_qualitative}. For intersection I in Fig.~\ref{fig:5_1_lujiazui}, most of the topology of the intersection is recovered by FlowMap, which is an impressive achievement without using HD maps. Moreover, even for HD map labeling, it is non-trivial to label human-like left turns for various road structures. On the other hand, it is notable that FlowMap provides human-like left turn paths. For example, the left turn marked in Fig.~\ref{fig:5_1_lujiazui} is very tight and the path should keep far away from the crosswalk. Utilizing traffic flow, FlowMap can generate such a kind of path naturally. For intersection II, it is challenging to generate the super long left turn and FlowMap has good results. We can conclude that FlowMap is good at generating human-like driving paths in open space, where road markings are unclear but driving patterns are clear.

\subsubsection{Map Error Detection}
As introduced in Sect.~\ref{sec:field_gen}, we can use the traffic flow field to detect potential map errors. A typical example is provided in Fig.~\ref{fig:4_2_xianshibiangeng}, where road construction occurred. The HD map is outdated for this example. In the conventional HD map maintenance pipeline, it is hard to quickly locate these road construction and re-routing events. Thanks to FlowMap, we can utilize traffic flow fields to identify these events. By comparing traffic flow results with HD map, mismatches can be easily identified and a map update signal can be sent out by FlowMap.

\subsection{Quantitative Results}
To quantitatively analyze the accuracy of FlowMap, we use reference paths from the HD map as a baseline. As shown in Fig.~\ref{fig:4_1_hd_compare}, we align the HD map with FlowMap and extract the corresponding reference path originating from a particular entry point. By comparing the path generated by FlowMap $l^z_\text{flow}$ to HD map reference path $l^z_\text{hd}$ , the following error metrics can be specified:
\begin{itemize}
\item Average displacement error (ADE): averaging over all channels (entry points) and flow candidates $\sum_{z} \sum_{l_\text{flow}\in\mathcal{L}_z} \text{AverageDisp}(l^z_\text{flow}, l^z_\text{hd})/ (\sum_z|\mathcal{L}_z|)$.
\item Maximum displacement error (MDE): averaging over maximum displacement error for each channel $\sum_{z} \sum_{l_\text{flow}\in\mathcal{L}_z} \text{MaxDisp}(l^z_\text{flow}, l^z_\text{hd})/ (\sum_z|\mathcal{L}_z|)$.
\item Displacement error at $x$ m (DE@$x$ m): averaging over all channels and flow candidates at a given distance $x$ m $\sum_{z} \sum_{l_\text{flow}\in\mathcal{L}_z} \text{DispAt}(l^z_\text{flow}, l^z_\text{hd}, x)/(\sum_z|\mathcal{L}_z|)$.
\end{itemize}
The results are given in TABLE.~\ref{tab:hd_compare}. It is notable that the HD map itself is not perfect, as shown in Fig.~\ref{fig:exp_hd}. But a statistically low error with respect to HD map reference can still represent that paths are generated correctly.

We further organize the results into different turn types. For \textit{inter-hard}, the ADE for going straight is $0.57$ m, which is very small and much lower than left-turn and right-turn. It means that the path generation algorithm is capable of generating high-quality paths. We argue that although path generation for left and right turns has higher errors, it does not necessarily mean worse quality. The errors may be induced by the feature that HD map reference is not human-like for left and right turns, as shown in Fig.~\ref{fig:4_1_hd_compare}. The standard deviation for left and right turns is significantly larger than going straight since human driving behaviors may differ when taking turns. In general, errors for \textit{inter-hard} are larger than \textit{inter-mild} except for right turns. The ``-'' in the table means that not enough candidates for this length.

\section{Conclusion and Future Work}\label{sec:conclusion}

In this paper, we propose FlowMap, a lightweight path generation framework for automated vehicles using traffic flow. The framewok can work in open space where there are no ``roads'' from sensor readings, by utilizing traffic flow information. To facilitate path planning, traffic flow fields are synthesized. Based on traffic flow fields, a path generation algorithm is proposed to extract human-like drivable paths. The whole framework is extensively evaluated using real-world driving data.  

\bibliography{paper}
\end{document}